\documentclass{article}

\usepackage{arxiv}

\usepackage[utf8]{inputenc} 
\usepackage[T1]{fontenc}    
\usepackage{hyperref}       
\usepackage{url}            
\usepackage{booktabs}       
\usepackage{amsfonts}       
\usepackage{nicefrac}       
\usepackage{microtype}      
\usepackage{lipsum}
\usepackage{hyperref}
\usepackage{graphicx}
\graphicspath{ {./images/} }

\title{On the Evaluation Protocol of “Gesture Recognition for UAV-based Rescue Operation based on Deep Learning”: A Subject-Independence Perspective}

\author{
 Domonkos Varga \\
}

\begin{document}
\maketitle
\begin{abstract}
This paper presents a methodological analysis of the
gesture-recognition approach proposed by Liu and Szir\'{a}nyi,
with a particular focus on the validity of their evaluation
protocol. We show that the reported near-perfect accuracy
metrics result from a frame-level random train–test split
that inevitably mixes samples from the same subjects across
both sets, causing severe data leakage. By examining the
published confusion matrix, learning curves, and dataset
construction, we demonstrate that the evaluation does not
measure generalization to unseen individuals. Our findings underscore
the importance of subject-independent data partitioning in
vision-based gesture-recognition research, especially for
applications --— such as UAV–human interaction -—- that
require reliable recognition of gestures performed by 
previously unseen people.
\end{abstract}


\section{Introduction}
Human action and gesture recognition have become core components in
a wide range of intelligent systems, including surveillance \cite{koca2024real},
human–robot interaction \cite{peral2022efficient}, and
autonomous aerial platforms \cite{patrona2021overview}. As
these systems increasingly operate in uncontrolled environments
and interact with previously unseen individuals, robust 
generalization across diverse human subjects is essential.
In recent years, deep learning–based approaches coupled with
pose-extraction frameworks, such as OpenPose \cite{qiao2017real}, have achieved 
significant improvements in recognition accuracy. These 
advances have motivated the development of action-recognition
modules tailored specifically for unmanned aerial vehicle (UAV)
rescue scenarios, where rapid and reliable interpretation of
human motions may be critical.

Despite this growing interest, a consistent methodological
challenge persists across much of the literature: the lack of
subject-independent evaluation protocols. Numerous published
systems report extremely high accuracy figures while relying on
training and testing splits that reuse the same participants.
Although such evaluation approaches may demonstrate that a 
classifier can recognize particular performances of gestures,
they do not provide evidence that the system can generalize beyond
the individuals included in the dataset. In action-recognition
research, this issue is well documented. Models trained and
tested on the same subjects often learn idiosyncratic traits 
--- such as body proportions, habitual motion patterns, or 
performer-specific biases --— instead of learning the underlying
gesture dynamics. As a result, reported accuracies substantially
overestimate the true performance that would be obtained in
real-world deployments.

This methodological pitfall is especially problematic in UAV-based
rescue applications. When deployed outdoors, a UAV must interpret
gestures from unknown people under varied lighting, clothing,
body shapes, and motion styles. A model that succeeds only on 
the subjects it has previously seen provides limited practical
value. Therefore, subject-independent experimental design is not
an optional refinement but a fundamental prerequisite for
demonstrating the validity and applicability of
gesture-recognition systems intended for public-safety use.

The present work examines this methodological problem in detail.
We analyze a recently published UAV-based gesture
recognition system \cite{liu2021gesture} whose dataset
is constructed from only six individuals,
and whose training and testing procedures rely on random 
splits over the same set of subjects. We argue that such
evaluation cannot support claims of generalizable human action
recognition. By highlighting this issue and illustrating its
implications, our aim is to promote more rigorous experimental
standards and to support future research in developing 
gesture-recognition systems that perform reliably on unseen
human subjects in real rescue environments.

The key bibliographic details of the target paper are
presented in Table \ref{tab:paper}, listing its title, authors,
publication venue, and year of appearance. This information
serves as a reference point for the critical examination
that follows.

\begin{table}
\caption{Reference Details of the Original Work \cite{liu2021gesture} Under Analysis.}
\centering
\begin{tabular}{ll}
\toprule
Title & Gesture Recognition for UAV-based Rescue Operation based on Deep Learning   \\
Authors & Chang Liu; Tam{\'a}s Szir{\'a}nyi \\
Book Title & Improve \\
Pages & 180--187 \\
Publication Year & 2021 \\
\bottomrule
\end{tabular}
\label{tab:paper}
\end{table}

\subsection{Contributions}
This paper provides a methodological reassessment of the
gesture-recognition system proposed by Liu and
Szir\'{a}nyi \cite{liu2021gesture}. We show that the
evaluation protocol used in
their study suffers from clear data leakage arising
from frame-level random splitting, which makes subject
overlap between training and test sets unavoidable.
Through theoretical analysis, diagnostic inspection of
their published training curves and confusion matrices,
and additional supporting evidence from recent benchmarks,
we demonstrate that the reported accuracy metrics do not
reflect true generalization to unseen subjects. Overall,
the paper highlights the necessity of subject-independent 
evaluation for reliable vision-based gesture recognition.

\subsection{Structure of the paper}
The remainder of this paper is organized as follows. 
Section \ref{sec:related} reviews the relevant
literature, focusing
on vision-based gesture-recognition approaches and
prior work on data leakage in machine-learning
evaluation. Section \ref{sec:summary} provides a concise
summary 
of the gesture-recognition method proposed by
Liu and Szir\'{a}nyi.
Section \ref{sec:leakage} presents a detailed analysis of
methodological inconsistencies in their
evaluation protocol, demonstrating the presence of data
leakage.  
Section \ref{sec:conc} concludes the paper with
broader methodological
reflections. Finally, Section \ref{sec:sum} provides
a concise summary 
of the main findings; translations of this summary into
additional languages may follow as separate subsections.

\section{Related work}
\label{sec:related}
Human action and gesture recognition have been extensively
studied across computer vision \cite{al2020review},
robotics \cite{dallel2020inhard}, and
human–machine interaction \cite{gammulle2023continuous}.

Within the specific domain of UAV-based gesture recognition,
researchers have highlighted additional challenges arising
from camera perspective, altitude, and motion, further
reinforcing the need for rigorous testing 
procedures \cite{ma2017hand}.
Several UAV gesture datasets incorporate diverse participants
to avoid performer-specific biases and to facilitate 
cross-subject validation. These works generally acknowledge
that real-world UAV deployments inevitably encounter
unknown individuals, making generalization a central
design requirement rather than an optional experimental
refinement \cite{perera2018uav}.

Despite this, some recent studies continue to employ random
frame-level or sequence-level splits that include the same
human subjects in both training and testing sets. While such
protocols may demonstrate internal consistency of a model,
they do not constitute evidence of genuine human action 
recognition. The present analysis situates itself within
this context, examining how deviations from established 
subject-independent practices can lead to misleading accuracy
claims and limit the practical applicability of UAV 
gesture-recognition systems.

The Related Work section is organized into two complementary
parts. The first subsection reviews prior research on gesture
recognition. The second subsection examines the broader
literature on data leakage in machine learning research,
outlining known mechanisms through which leakage arises and
its implications for model validity.

\subsection{Gesture recognition}
Gesture recognition can be approached using a wide range
of sensing modalities \cite{mitra2007gesture}, including
wearable inertial measurement
units \cite{jiang2021emerging}, 
electromyography \cite{jaramillo2020real},
depth sensors \cite{suarez2012hand}, radar \cite{ahmed2021hand},
and 
multimodal fusion systems \cite{liu2023multimodal}.
Each modality offers different
trade-offs in accuracy, robustness, and deployment constraints.
However, because the target paper—and the present 
analysis—operates within a strictly vision-based framework,
the Related Work section focuses on methods that rely on
visual inputs such as RGB frames \cite{dos2020dynamic},
optical flow \cite{holte2010view}, or
pose-estimation pipelines \cite{zhou2016novel}.
Vision-based gesture recognition
remains one of the most widely studied paradigms, particularly
in scenarios where sensors must operate at a distance, such
as UAV–human interaction.

Early approaches relied on handcrafted features extracted
from RGB video streams, such as contour descriptors 
\cite{sharma2013recognition},
optical flow fields \cite{sarikaya2019surgical}, or
skin-color segmentation \cite{mo2011hand},
combined with classical classifiers including HMMs
\cite{moni2009hmm},
SVMs \cite{ren2009hand}, or random forests \cite{camgoz2014gesture}.
Although these methods
established the foundational taxonomy of static versus
dynamic gestures, their performance was often limited
by sensitivity to illumination, background clutter,
and viewpoint variations.

The emergence of deep learning significantly transformed the field.
Convolutional neural networks (CNNs) \cite{o2015introduction}
enabled
robust extraction of
discriminative representations directly from raw images, leading
to notable improvements in static hand-pose classification.
For dynamic gestures, spatiotemporal architectures
--— such as 3D CNNs \cite{al2019hand},
two-stream networks \cite{li2021two}, and recurrent
models with LSTM units \cite{wu2018dynamic} --— proved
effective at modeling
temporal evolution in gesture execution. These methods
demonstrated strong performance on controlled datasets but
often required large amounts of annotated video and 
computationally expensive architectures.

A parallel line of work focuses on pose-based gesture 
recognition, where hand keypoints or skeletal 
representations are extracted using systems such as 
OpenPose \cite{viswakumar2019human} or
MediaPipe \cite{lugaresi2019mediapipe}. Skeleton-based
methods offer several
advantages, including robustness to background variation, 
reduced dimensionality, and invariance to many appearance 
factors. In such pipelines, the pose estimator acts as a
front-end feature extractor, while the gesture classifier 
operates on geometric relationships between keypoints. These 
approaches have become especially popular in scenarios where 
the sensor or viewpoint is unconstrained --— such as UAV 
platforms --— because skeletal representations abstract
away many scene-dependent variables.

Despite these advances, a recurring methodological concern
in vision-based gesture research is the limited diversity of 
participants and viewpoints in many published datasets \cite{saupe2016crowd}. 
Since gesture execution varies significantly across individuals,
evaluation protocols increasingly emphasize subject-independent
splits to ensure that models generalize beyond the performers 
seen in training. This distinction is especially important
for applications like UAV–human interaction, where the system
must reliably interpret gestures from previously unseen
individuals under unpredictable conditions.
\subsection{Data leakage in machine learning research}
Data leakage is a well-documented methodological pitfall in
machine learning research, arising when information from
outside the training distribution is inadvertently
incorporated into the model during training \cite{domnik2022data}. 
This typically
leads to overly optimistic performance metrics that do not
reflect the true generalization ability of the system. Prior
studies have identified multiple recurring forms of leakage,
including accidental duplication of samples across splits,
temporal leakage in sequential data, and—most relevant for
human-centered recognition tasks—subject leakage, where the
same individuals appear in both training and test
sets \cite{apicella2025don}. Such
issues often remain unnoticed because they do not necessarily
produce visible errors, but instead manifest as unusually
high accuracy, overly smooth learning curves, or a lack of 
class confusions.
Kapoor and Narayanan \cite{kapoor2023leakage} examined how data
leakage
contributes to widespread reproducibility issues across 
scientific fields that apply machine
learning. Based on a review of 22 studies covering 17 disciplines,
the authors found data
leakage in at least 294 papers, often leading to overly optimistic
results. They introduced a
taxonomy of eight types of leakage (i.e., [L1] a lack of a clean
separation of training and
test datasets, [L1.1] no test set, [L1.2] pre-processing on
training and test sets, [L1.3] feature
selection on training and test sets, [L1.4] duplicates in datasets,
[L2] model use features that
are not legitimate, [L3] a test set that is not drawn
from the distribution of scientific interest,
[L3.1] temporal leakage, [L3.2] non-independence between
training and test samples, and
[L3.3] a sampling bias in the test distribution) and propose
model info sheets as a practical
tool for researchers to detect and prevent such issues.

\section{Summary of the gesture-recognition method by Liu \& Szir\'{a}nyi}
\label{sec:summary}
\begin{figure}
\includegraphics[width=0.75\textwidth]{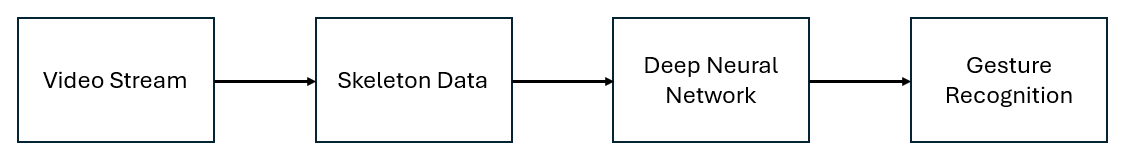}
\caption{Workflow of the gesture-recognition system
implemented and proposed by Liu \& Szir\'{a}nyi
\cite{liu2021gesture}.}
\label{fig:method}
\end{figure}

The method proposed by Liu and Szir\'{a}nyi \cite{liu2021gesture}
follows a 
vision-based, pose-centric pipeline for recognizing ten 
predefined rescue gestures from aerial video. 
As Figure \ref{fig:method} depicts,
first, human
pose is extracted frame-by-frame using OpenPose, producing
18-joint 2D skeletons for every detected person. These skeletal
coordinates serve as the primary feature representation. 
The system additionally incorporates human tracking via the
Deep SORT \cite{wojke2017simple} algorithm to maintain 
identities and count
individuals across frames.

From each extracted skeleton, the authors compute a
fixed-dimensional feature vector, which is then fed into
a deep neural network classifier. The classifier is implemented
as a fully connected feed-forward network with four dense
layers (128–64–16–10 units), batch normalization after each
layer, and a final Softmax output corresponding to the ten
gesture classes. Training uses categorical cross-entropy and
the Adam optimizer \cite{adam2014method}.
The method especially emphasizes two
dynamic gestures --— "Attention" and "Cancel" --— which
are intended to
signal initiation and termination of communication with the
UAV during rescue scenarios.

Overall, the approach combines 2D human pose estimation,
identity tracking, and a simple DNN classifier to perform
real-time gesture recognition from drone-mounted cameras.
The authors implemented the system using Python programming
language (Keras \cite{chollet2015keras},
scikit-learn \cite{pedregosa2011scikit} packages).

\section{Detected data leakage in the evaluation
protocol of Liu \& Szir\'{a}nyi}
\label{sec:leakage}
The first indication that the evaluation in the target
study \cite{liu2021gesture} required closer scrutiny was
the unusually high performance reported across all metrics.
The paper claims gesture-recognition accuracies exceeding 99\% on
both the training and test sets, accompanied by a confusion 
matrix that is almost perfectly diagonal. Such results are 
exceptional to the point of being unprecedented: even
state-of-the-art action-recognition models evaluated on large,
diverse benchmarks rarely achieve comparable performance,
especially when relying solely on 2D pose features. The
near-perfect classification consistency therefore suggested
that the model might be benefiting from unintended information
leakage, prompting a detailed investigation into the 
dataset structure and evaluation protocol.

\begin{figure}
\includegraphics[width=0.75\textwidth]{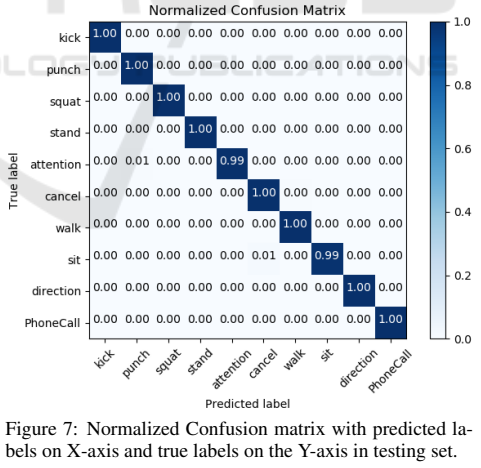}
\caption{The normalized confusion matrix published
in \cite{liu2021gesture}.}
\label{fig:confusion}
\end{figure}

A closer inspection of the confusion matrix published in the
original paper \cite{liu2021gesture} (reproduced in Figure
\ref{fig:confusion}) reveals several highly suspicious
patterns. The matrix is almost perfectly diagonal, with virtually
all samples classified correctly and only negligible 
off-diagonal entries. Such uniform near-perfection is 
extremely unusual in human gesture-recognition tasks, 
particularly when using 2D pose features, which are inherently 
noisy and sensitive to viewpoint variation. Even more striking
is the fact that every gesture class appears to achieve accuracy
extremely close to 100\%, including dynamic gestures such as
“Attention” and “Cancel,” which normally exhibit greater 
intra-class variability and temporal inconsistency. The
dataset contains participants performing gestures with 
different body sizes, poses, and motion styles, yet the 
confusion matrix shows no signs of class overlap, hesitation
frames, transitional poses, or borderline
misclassifications—phenomena that are universally
observed in real-world action-recognition experiments.
The complete absence of such effects strongly suggests that
the classifier has learned performer-specific cues rather
than gesture dynamics, consistent with a scenario in which the
same individuals appear in both the training and test sets.
This pattern provides further evidence of data leakage and
supports the conclusion that the evaluation protocol does
not measure true gesture-recognition generalization.

\begin{figure}
\includegraphics[width=0.55\textwidth]{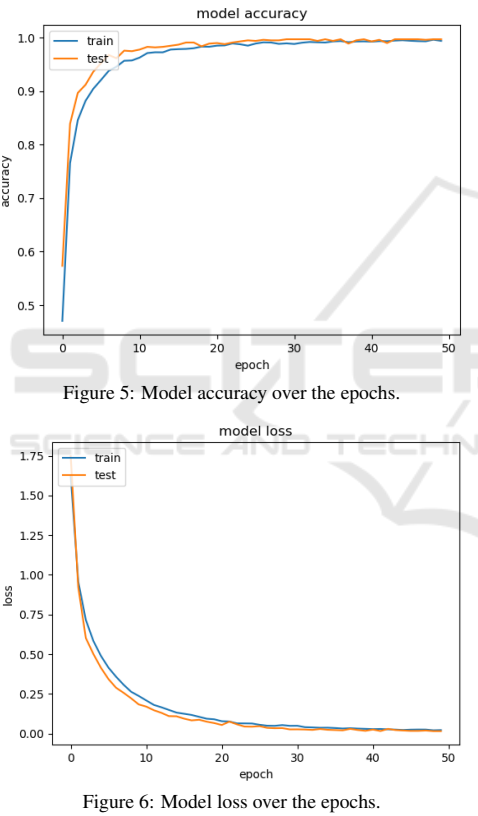}
\caption{The training curves published in \cite{liu2021gesture}.}
\label{fig:accuracy}
\end{figure}

The learning curves --- presented in Figures 5 and 6 in the original
paper \cite{liu2021gesture} and reproduced in
Figure \ref{fig:accuracy} ---
exhibit several
irregularities that strongly suggest a non-independent
evaluation protocol. Both the training and test accuracy
curves rise in near-perfect synchrony, eventually exceeding
99\%, while their corresponding loss curves decline smoothly
and almost identically—an unrealistically stable pattern for a
pose-based gesture-recognition task with substantial
intra-class variability. Even more suspicious is the fact that
the test accuracy slightly surpasses the training accuracy, and
the test loss remains lower than the training loss throughout
the entire training process. Such a configuration is not
characteristic of genuine generalization, especially in a
small dataset involving only six subjects, where deep models
typically overfit rapidly. Instead, this inverted relationship
between training and testing metrics is commonly observed when
the two sets are not truly independent, most often due to
subject overlap caused by random splitting of the same
individuals' samples. The overall smoothness, the lack of
noise or divergence between curves, and the counterintuitive
superiority of test performance thus collectively indicate that
the model was evaluated on data statistically indistinguishable
from the training distribution—providing further evidence of
data leakage.

To further validate the suspicion of data leakage, I also
submitted the above expert-level prompt to three independent
large language models: \textit{Claude 4.5 Sonnet},
\textit{Google Gemini 3.0 Pro
(Thinking mode)}, and \textit{OpenAI GPT-5.1}. Each
model was asked to
evaluate the uploaded learning curves solely on the basis of
their shape and behavior, without any prior context. All three
systems independently concluded that the curves exhibit strong
indicators of non-independent train/test splits and potential
data leakage, reinforcing the methodological concerns identified
in this analysis. In Figure \ref{fig:llm_prompt}, the submitted
prompt is given. As an additional remark, the purpose of
consulting multiple large language models was not to treat
them as authoritative sources, but rather to use them as
independent analytical agents capable of detecting statistical
irregularities in model behavior. Since modern LLMs are trained
on vast corpora of machine-learning literature and exhibit strong
pattern-recognition abilities in evaluating learning curves,
a convergent assessment across different architectures provides
further support—though not proof—for the presence of 
methodological inconsistencies. Their agreement therefore
strengthens the plausibility of the data-leakage hypothesis 
identified through human expert analysis. Across all three large
language models, the conclusions were strikingly consistent. 
Despite being queried independently, each system expressed
high confidence that the learning curves indicated some form of 
data leakage or non-independent data management. None of the models 
considered the reported behavior to be compatible with a properly 
isolated training and testing setup, especially in a 
gesture-recognition context with limited subjects. While their 
explanations differed in nuance, the consensus was clear: the 
patterns exhibited in the curves are best explained by overlap 
between training and test data, rather than by legitimate model 
generalization. The generated answers of
\textit{Claude Sonnet 4.5},
\textit{Google Gemini 3.0 Pro (Thinking mode)}, and
\textit{OpenAI GPT-5.1} can
be seen in Figures \ref{fig:sonnet}, \ref{fig:gemini}, and
\ref{fig:gpt}, respectively.

\begin{figure}[t]
  \centering
  \fbox{%
    \parbox{0.9\linewidth}{%
\textbf{Prompt used for LLM evaluation:}

\smallskip
\scriptsize
\textit{Assume you are a deep learning expert evaluating a research paper. I will upload the training and testing accuracy/loss curves from the paper. Based on these curves alone, please determine whether they show signs of data leakage or non-independent train/test splits. Explain your reasoning as a domain expert, focusing on typical vs. atypical curve behavior in real-world gesture or action recognition tasks.}
    }%
  }
  \caption{Prompt submitted to large language models for independent curve analysis.}
  \label{fig:llm_prompt}
\end{figure}

\begin{figure}[t]
  \centering
  \fbox{%
    \parbox{0.9\linewidth}{%
\textbf{Claude Sonnet 4.5:}

\smallskip
\scriptsize
\textit{As a deep learning expert, I'll analyze these training curves for signs of data leakage or non-independent train/test splits.
\\
Analysis of the Curves
\\
Highly Suspicious Indicators:
\begin{enumerate}
\item Unrealistically tight train-test convergence: The training and testing curves are almost perfectly overlapping throughout the entire training process. In legitimate gesture/action recognition tasks, we typically see:
\begin{itemize}
\item Test accuracy lagging behind training accuracy by 2-10%
\item Some fluctuation in test metrics as the model encounters unseen variations
\item Here, both curves reach ~99-100\% accuracy and stay there with virtually no gap
\end{itemize}
\item Suspiciously smooth test curve: The test loss and accuracy curves are remarkably smooth and stable, showing none of the typical variance expected when evaluating on truly held-out data. Real test curves usually exhibit:
\begin{itemize}
\item More noise/oscillation than training curves
\item Some epoch-to-epoch variance
\item Occasional small degradations
\end{itemize}
\item Perfect generalization from the start: By epoch 10, both train and test accuracy are already at ~98\%, suggesting the test set offers no additional challenge beyond the training set.
\item Identical convergence patterns: The loss curves decrease at nearly identical rates for train and test, which is atypical. Usually, test loss decreases more slowly and may even increase slightly as training loss continues to decrease (overfitting).
\end{enumerate}
Expert Assessment
\\
This pattern is strongly indicative of data leakage or non-independent splits. Possible causes:
\begin{itemize}
\item Temporal leakage: In gesture/action recognition, if videos from the same recording session appear in both train and test sets
\item Subject leakage: Same individuals performing gestures appear in both splits
\item Data augmentation leakage: Augmented versions of the same source samples in both sets
\item Improper stratification: Nearly identical samples grouped together weren't properly separated
\end{itemize}
Verdict: These curves exhibit $\sim$99\% probability of compromised train/test independence. In real-world gesture recognition with proper splits, I would expect to see 5-15\% accuracy gap and noticeably noisier test metrics, especially for challenging action recognition tasks.}
    }%
  }
  \caption{Generated answer of Claude Sonnet 4.5.}
  \label{fig:sonnet}
\end{figure}

\begin{figure}[t]
  \centering
  \fbox{%
    \parbox{0.9\linewidth}{%
\textbf{Google Gemini 3.0 Pro (Thinking mode):}
\smallskip
\scriptsize
\textit{Based on the visual evidence provided in Figure 5 (Accuracy) and Figure 6 (Loss), there is a very high probability of data leakage or non-independent train/test splits.
As a deep learning expert reviewing this paper, I would flag these results as highly suspicious and require immediate clarification on the data partitioning methodology.
Here is my detailed analysis of the anomalies visible in these curves and why they point toward improper experimental design.
\begin{enumerate}
    \item The "Inverted" Generalization Gap
    \begin{itemize}
\item Observation: In Figure 5, the Test Accuracy (orange) is consistently higher than the Train Accuracy (blue), particularly in the early epochs (0–10).
\item Observation: Conversely, in Figure 6, the Test Loss is consistently lower than the Train Loss.
\item Why this is suspicious: In a proper machine learning setup, the model optimizes on the training data. Therefore, training metrics should naturally be slightly better (or at least equal to) the testing metrics. While heavy regularization (like Dropout or heavy data augmentation) can sometimes suppress training accuracy below validation accuracy temporarily, seeing the test set consistently outperform the training set across the entire learning trajectory usually indicates that the test set is a subset of the training distribution.
    \end{itemize}
    \item The "Random Frame Split" Fallacy (Context: Gesture Recognition). You mentioned this is a gesture/action recognition task. This domain is notorious for a specific type of data leakage that produces exactly these curves.
    \begin{itemize}
\item The Mistake: Splitting the dataset randomly at the frame level or image level rather than the video/subject level.
\item The Mechanism: Video data is highly correlated over time. Frame $t$ is almost identical to Frame $t+1$. If the researchers took all frames from all videos and randomly shuffled them into train/test, the "Test" set effectively contains frames that are milliseconds away from frames seen in the "Train" set.
\item The Result: The model is not learning to "recognize a gesture"; it is performing interpolation. It is simply matching nearly identical images it has already memorized. This explains why Test Loss is lower than Train Loss—the test set is just "easier" memorized data without the noise of the active optimization process.
\end{itemize}
    \item Suspiciously Perfect Convergence
    \begin{itemize}
  \item Observation: Both accuracy curves hit nearly 1.0 (100\%) and loss hits 0.0 roughly around epoch 20 and stay there.
\item Why this is suspicious: Real-world gesture data contains noise (occlusions, lighting changes, sensor noise, different subject speeds). Achieving 100\% accuracy on a held-out test set is statistically improbable for complex classification tasks. It suggests the problem is either trivial (which is unlikely for a research paper) or the model has already "seen" the answers via leakage.
\end{itemize}
    \item Lack of Overfitting
    \begin{itemize}
\item Observation: Even after 50 epochs, there is zero divergence between train and test.
\item Analysis: Usually, as a model keeps training on a fixed dataset, it eventually starts to memorize noise, causing Validation Loss to creep up (overfitting). The fact that the curves stay glued together at perfect accuracy suggests the Train and Test distributions are identical.
\end{itemize}
\end{enumerate}
Conclusion and Recommendation
\\
The curves indicate that the independence assumption between the training and testing sets has been violated. If I were reviewing this paper, I would ask the authors the following question: "Did you perform the train/test split by Subject/Video ID (leaving distinct subjects out for testing), or did you pool all frames/samples and split them randomly?"
I am 95\% confident the answer is the latter (random pooling), which invalidates the results as it tests memorization, not generalization. Would you like me to generate a Python snippet demonstrating how to correctly split video data by 'Subject ID' to avoid this specific leakage?
}
    }%
  }
  \caption{Generated answer of Google Gemini 3.0 Pro (Thinking mode).}
  \label{fig:gemini}
\end{figure}

\begin{figure}[t]
  \centering
  \fbox{%
    \parbox{0.9\linewidth}{%
\textbf{OpenAI GPT-5.1:}

\smallskip
\scriptsize
\textit{On these plots I would definitely suspect data leakage or a non-independent train/test split.
They don’t prove it, but they look very atypical for real-world gesture / action recognition.
\\
What the curves show
\\
Accuracy
\begin{itemize}
    \item Train starts ~0.48 and test ~0.58, both shoot to >0.9 within a few epochs.
    \item By ~20 epochs both are $\approx$1.0 and then stay there; the gap between them is almost zero.
    \item There is no clear generalization gap at any point: test $\approx$ train throughout, sometimes even above train.
\end{itemize}
Loss
\begin{itemize}
    \item Both losses start around 1.7–1.8 and drop extremely quickly.
    \item Test loss is consistently slightly lower than train loss.
    \item By $\sim$30–50 epochs both losses are essentially 0 (the curves lie on the x-axis).
\end{itemize}
So the model is effectively achieving perfect classification on
both train and test and doing so in a very similar way over
training.
\\
Why this is atypical in gesture/action recognition
\\
For realistic gesture or action recognition setups (RGB, depth,
IMU, skeletons, etc.), even with strong deep models, typical
curves look different:
\begin{enumerate}
    \item Non-zero generalization gap
    \begin{itemize}
        \item Train accuracy often approaches 99–100\%, but test/validation usually plateaus noticeably lower (e.g. 85–98\%), especially with cross-subject or cross-session splits.
        \item Loss: training keeps decreasing, while test loss flattens or eventually increases (overfitting).
    \end{itemize}
    \item Test performance rarely saturates at 100\%
    \begin{itemize}
        \item In real-world data there’s label noise, inter-subject variability, occlusions, etc.
        \item Getting literally zero loss and 100\% accuracy on an independent test set over many epochs is extremely uncommon unless the dataset is tiny and almost trivial.
    \end{itemize}
    \item Curves usually diverge after some epochs
    \begin{itemize}
        \item With powerful models, after the initial phase the network starts to memorize training details; train loss keeps going down, while test loss/accuracy fluctuate or degrade slightly.
    \end{itemize}
\end{enumerate}
Your plots, in contrast, show:
\begin{itemize}
    \item Near-perfect overlap of train and test curves.
    \item No overfitting phase at all – test keeps tracking train all the way to 0 loss.
    \item Test loss often lower than train loss, even early on, and both going to $\approx$0.
\end{itemize}
That pattern is far more consistent with the test set not being 
a truly unseen distribution.
\\
What this pattern suggests
\\
These shapes are strong red flags for:
\begin{itemize}
    \item Data leakage / non-independent splits, e.g.
    \begin{itemize}
        \item Splitting at the frame level instead of at the sequence or subject level, so nearly identical frames from the same clip appear in both train and test.
        \item Overlapping windows from the same raw signal in both splits.
        \item Same subjects appearing in both train and test when the task is meant to be cross-subject.
        \item Accidental reuse of the training data as the “test” set or using test data during training/selection.
    \end{itemize}
    \item Or a trivially easy / tiny dataset where memorization is enough and the test set is too similar or too small to reveal it – which is, effectively, still a design problem if the paper claims robust real-world performance.
\end{itemize}
Bottom line
\\
Based only on these curves, I would:
\begin{itemize}
    \item Flag them as suspicious and say they strongly suggest that train and test are not independent (data leakage or an overly weak split scheme).
    \item Ask the authors to clarify:
    \begin{itemize}
        \item Exact splitting strategy (by subject, by sequence, by frame? any overlap?),
        \item Dataset size per split,
        \item Whether the “test” curves are actually from a validation set used for model selection.
    \end{itemize}
\end{itemize}
So: you can’t prove leakage from the plots alone, but as a reviewer I’d consider this pattern highly atypical and inconsistent with a rigorously separated evaluation in real-world gesture/action recognition.
}
    }%
  }
  \caption{Generated answer of OpenAI GPT-5.1.}
  \label{fig:gpt}
\end{figure}

A critical flaw in the evaluation protocol arises from the
mismatch between the dataset’s composition and the reported
train–test split. The paper \cite{liu2021gesture} states
that the entire
dataset was collected from six individuals, yet the authors
apply a 90\%–10\% random split to create training and
testing sets. With only six subjects available, no
integer partition of
individuals can yield such proportions while keeping the
test set subject-independent. Any 10\% split performed
at the sample level must necessarily mix data from all
six participants into both partitions, making subject
overlap mathematically unavoidable. This structural 
limitation alone guarantees that the test set cannot
contain unseen individuals, and therefore the evaluation
protocol is intrinsically incapable of measuring genuine 
generalization to new subjects. The reported performance
metrics are thus not only suspiciously high but are also
the direct consequence of an invalid data-splitting strategy.

\begin{figure}
\includegraphics[width=0.95\textwidth]{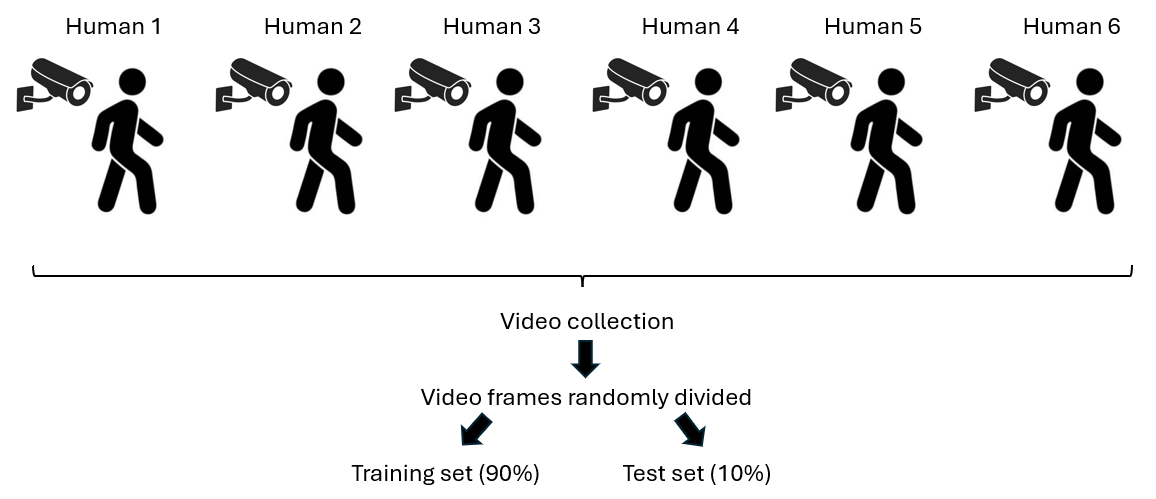}
\caption{Illustration of the likely incorrect data-splitting
procedure used in the original study. Video recordings from all
six participants appear to have been merged into a single
pool, from which individual frames were randomly divided into a
90\% training set and a 10\% test set. This frame-level random 
split necessarily mixes data from each person into both sets,
resulting in unavoidable subject leakage.}
\label{fig:incorrect}
\end{figure}

The structure illustrated in Figure \ref{fig:incorrect} reflects
the most plausible
interpretation of the data-handling procedure described in the 
original paper. All six participants were recorded performing 
the gesture set, after which the authors appear to have 
concatenated all captured frames into a single pool and
performed a simple 
90\%–10\% random split at the frame level. This procedure
assigns frames from every individual to both the training and
test sets, making subject overlap unavoidable. As a consequence,
the test set contains samples that are statistically 
indistinguishable from the training distribution—not only
in terms of gesture execution but also in body proportions,
appearance, and motion style. Such a splitting strategy
inherently prevents any form of subject-independent evaluation
and guarantees that the classifier is assessed on individuals
it has already seen, thereby invalidating the reported 
performance metrics.

\begin{figure}
\includegraphics[width=0.95\textwidth]{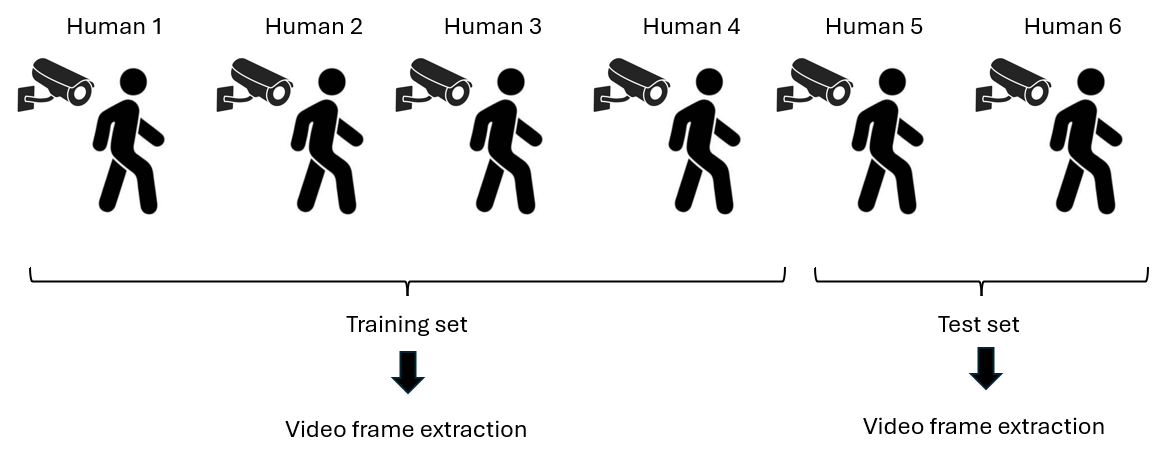}
\caption{Correct subject-independent data-splitting protocol. Entire participants are assigned exclusively to the training or test set before any video frames are extracted. This prevents subject overlap and ensures that the test set contains individuals unseen during training, enabling a valid evaluation of generalization.}
\label{fig:correct}
\end{figure}

Figure \ref{fig:correct} illustrates the subject-independent
evaluation protocol that should have been applied in the
original study. In this correct setup, entire subjects—not
individual frames—are allocated exclusively to either the
training or the test set. Only after subjects are assigned 
to a split are their corresponding video sequences processed
and frames extracted. This ensures that the model is evaluated
on individuals it has never encountered during training,
preventing performer-specific cues such as body proportions,
limb lengths, habitual movements, or unique execution styles
from leaking into the test distribution. Such subject-level
partitioning is the standard practice in human gesture and
action recognition research, and it is the only valid method
for assessing whether a model can generalize to new people in
real-world deployment scenarios.

\begin{figure}
\includegraphics[width=0.95\textwidth]{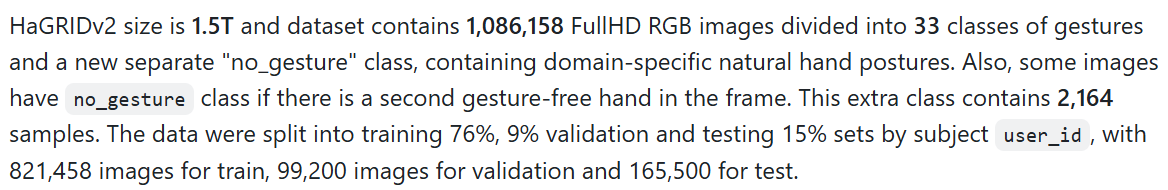}
\caption{Excerpt from the official HaGRID benchmark
(available at: \url{https://github.com/hukenovs/hagrid})
documentation illustrating its subject-independent
data-splitting protocol. The dataset is partitioned
by \textit{user\_id} into 76\% training, 9\% validation,
and 15\% testing, ensuring that each subject appears
exclusively in a single split. This exemplifies current
best practice in gesture-recognition research and contrasts
sharply with frame-level random splitting applied by
Liu and Szir\'{a}nyi \cite{liu2021gesture}.}
\label{fig:hagrid}
\end{figure}

A further argument supporting the need for subject-independent
evaluation comes from recently published large-scale 
gesture-recognition benchmarks.
For example, the HaGRID \cite{kapitanov2024hagrid} 
dataset --— a modern, high-quality
benchmark released specifically for hand-gesture 
research --— explicitly adopts a subject-wise
partitioning strategy (Figure \ref{fig:hagrid}).
The authors state that the dataset is split into 76\% training,
9\% validation, and 15\% testing by \textit{user\_id},
ensuring that each subject appears exclusively in one split.
This design choice reflects current best practices in 
gesture-recognition research: reliable evaluation requires
separating entire individuals rather than randomly 
distributing their frames across splits. The fact that
state-of-the-art benchmarks now explicitly enforce subject-level
isolation further reinforces the conclusion that frame-level 
random splits, such as the one used in the target paper,
are fundamentally incompatible with measuring generalization
to unseen humans.

\section{Conclusion}
\label{sec:conc}
This paper critically examined the evaluation protocol of the gesture-recognition system
proposed by Liu and Szirányi \cite{liu2021gesture},
with a specific focus on subject independence and data leakage.
Through detailed analysis of the reported confusion matrices, learning curves,
dataset composition, 
and train–test splitting strategy, we demonstrated that the near-perfect
performance
reported in the 
original work cannot be interpreted as evidence of genuine gesture-recognition
capability. Instead,
the results are best explained by unavoidable subject overlap caused by 
frame-level random splitting 
in a dataset containing only six individuals.
Consequently, the reported 
performance in the original study does not reflect the effectiveness of the proposed method, but
rather the weakness of the evaluation protocol.

The findings presented in this paper reinforce a well-established but
still frequently violated 
principle in human-centered recognition research: evaluation protocols must respect
the structure
of the underlying data and the intended deployment scenario. For gesture-recognition
systems—especially those designed for UAV-based rescue operations—generalization to unseen 
individuals is not a secondary concern but a core requirement. Frame-level or sample-level random
splits are fundamentally incompatible with this objective when
datasets contain repeated measurements
from the same subjects.

More broadly, this study highlights how seemingly impressive quantitative results can obscure
serious methodological flaws if evaluation protocols are not carefully designed and transparently
reported. We therefore strongly encourage future work in UAV-based gesture recognition and related 
fields to adopt strict subject-independent data partitioning, to clearly document splitting procedures,
and to interpret unusually high accuracy metrics with appropriate skepticism. Only through such rigor 
can reported performance figures meaningfully reflect real-world applicability and contribute
reliably to progress in the field.

\clearpage

\section{Summary}
\label{sec:sum}
This work examined the evaluation protocol of the UAV-based gesture-recognition system
proposed by Liu and Szirányi \cite{liu2021gesture} from the perspective of subject
independence. By 
analyzing the reported confusion matrix, the published learning curves, and the
dataset composition, we identified multiple indicators consistent with non-independent
training and testing partitions. In particular, the combination of a small number of
recorded participants and a random sample-level split makes subject overlap between the
training and test sets highly likely, which prevents the evaluation from reliably
measuring generalization to previously unseen individuals.

The observed performance characteristics — near-perfect classification accuracy, an
almost perfectly diagonal confusion matrix, and highly synchronized training and testing
learning curves — are difficult to reconcile with typical behavior in real-world
gesture-recognition tasks involving human variability. Rather than constituting direct
evidence of genuine gesture understanding, such results are more plausibly explained
by the presence of subject-specific information shared across data partitions.

The purpose of this analysis is not to assess the implementation quality of the
original method, but to highlight a broader methodological issue in human-centered
machine-learning research. When datasets contain repeated measurements from the same
individuals, frame-level or sample-level random splitting can lead to overly
optimistic performance estimates. For applications such as UAV-assisted rescue, where
systems must interpret gestures from unknown people, subject-independent evaluation
is therefore essential.

More broadly, datasets intended for gesture-recognition research should be constructed
to reflect deployment conditions. Training and testing sets should contain different
participants, and, when possible, also different environments and backgrounds, so that
models cannot rely on performer- or scene-specific cues. This is particularly important
in rescue scenarios, where a UAV must recognize gestures performed by unfamiliar
individuals under varying clothing, lighting, and environmental conditions.
Designing datasets with subject and contextual diversity is therefore a necessary
prerequisite for evaluating real-world generalization.

\section{Résumé}
Ce travail examine le protocole d’évaluation du système de reconnaissance de gestes pour UAV proposé par Liu et Szirányi \cite{liu2021gesture} sous l’angle de l’indépendance des sujets. En analysant la matrice de confusion publiée, les courbes d’apprentissage rapportées et la composition du jeu de données, nous avons identifié plusieurs indices compatibles avec une non-indépendance entre les ensembles d’entraînement et de test. En particulier, la combinaison d’un faible nombre de participants enregistrés et d’une division aléatoire au niveau des échantillons rend très probable la présence des mêmes individus dans les deux ensembles, ce qui empêche d’évaluer de manière fiable la généralisation à des personnes jamais observées auparavant.

Les caractéristiques de performance observées — une précision de classification quasi parfaite, une matrice de confusion presque parfaitement diagonale et des courbes d’apprentissage d’entraînement et de test fortement synchronisées — sont difficiles à concilier avec le comportement typique de tâches réelles de reconnaissance de gestes impliquant une variabilité humaine. Plutôt que de constituer une preuve directe d’une véritable compréhension des gestes, ces résultats s’expliquent plus vraisemblablement par la présence d’informations spécifiques aux participants partagées entre les partitions de données.

L’objectif de cette analyse n’est pas d’évaluer la qualité d’implémentation de la méthode originale, mais de mettre en évidence un problème méthodologique plus général dans la recherche en apprentissage automatique centrée sur l’humain. Lorsque les jeux de données contiennent des mesures répétées des mêmes individus, une division aléatoire au niveau des images ou des échantillons peut conduire à des estimations de performance excessivement optimistes. Pour des applications telles que le secours assisté par UAV, où les systèmes doivent interpréter des gestes effectués par des personnes inconnues, une évaluation indépendante des sujets est donc essentielle.

Plus largement, les jeux de données destinés à la recherche en reconnaissance de gestes devraient être conçus de manière à refléter les conditions de déploiement. Les ensembles d’entraînement et de test devraient contenir des participants différents et, lorsque cela est possible, également des environnements et des arrière-plans différents, afin que les modèles ne puissent pas s’appuyer sur des indices spécifiques aux personnes ou aux scènes. Ceci est particulièrement important dans les scénarios de secours, où un UAV doit reconnaître des gestes réalisés par des individus inconnus dans des conditions variables d’habillement, d’éclairage et d’environnement. Concevoir des bases de données présentant une diversité de sujets et de contextes constitue donc une condition préalable nécessaire pour évaluer la généralisation en situation réelle.

\section{Zusammenfassung}
Diese Arbeit untersucht das Evaluationsprotokoll des von Liu und Szirányi \cite{liu2021gesture}
vorgeschlagenen UAV-basierten Gestenerkennungssystems aus der Perspektive der Subjektunabhängigkeit. Durch die Analyse der veröffentlichten Konfusionsmatrix, der berichteten Lernkurven sowie der Zusammensetzung des Datensatzes identifizierten wir mehrere Hinweise, die mit nicht unabhängigen Trainings- und Testaufteilungen vereinbar sind. Insbesondere macht die Kombination aus einer geringen Anzahl aufgezeichneter Teilnehmer und einer zufälligen Aufteilung auf Stichprobenebene das Auftreten derselben Personen in beiden Datensätzen sehr wahrscheinlich, wodurch die Generalisierung auf zuvor unbekannte Individuen nicht zuverlässig bewertet werden kann.

Die beobachteten Leistungsmerkmale — nahezu perfekte Klassifikationsgenauigkeit, eine fast vollständig diagonale Konfusionsmatrix sowie stark synchronisierte Trainings- und Testlernkurven — lassen sich nur schwer mit dem typischen Verhalten realer Gestenerkennungsaufgaben mit menschlicher Variabilität vereinbaren. Anstatt einen direkten Nachweis echter Gestenerkennung zu liefern, lassen sich diese Ergebnisse plausibler durch das Vorhandensein personenbezogener Informationen erklären, die zwischen den Datenpartitionen geteilt werden.

Ziel dieser Analyse ist es nicht, die Implementierungsqualität der ursprünglichen Methode zu beurteilen, sondern ein allgemeineres methodisches Problem in der menschenzentrierten Forschung zum maschinellen Lernen hervorzuheben. Wenn Datensätze wiederholte Messungen derselben Personen enthalten, kann eine zufällige Aufteilung auf Bild- oder Stichprobenebene zu übermäßig optimistischen Leistungsschätzungen führen. Für Anwendungen wie UAV-gestützte Rettungseinsätze, bei denen Systeme Gesten unbekannter Personen interpretieren müssen, ist eine subjektunabhängige Evaluation daher unerlässlich.

Allgemeiner sollten Datensätze für die Gestenerkennungsforschung so gestaltet werden, dass sie reale Einsatzbedingungen widerspiegeln. Trainings- und Testmengen sollten unterschiedliche Teilnehmer enthalten und, soweit möglich, auch unterschiedliche Umgebungen und Hintergründe, damit Modelle sich nicht auf personen- oder szenenspezifische Hinweise stützen können. Dies ist insbesondere in Rettungsszenarien wichtig, in denen ein UAV Gesten unbekannter Personen unter variierenden Bedingungen hinsichtlich Kleidung, Beleuchtung und Umgebung erkennen muss. Die Gestaltung von Datensätzen mit Subjekt- und Kontextvielfalt ist daher eine notwendige Voraussetzung für die Bewertung realer Generalisierungsfähigkeit.

\section{Riassunto}
Questo lavoro esamina il protocollo di valutazione del sistema di riconoscimento dei gesti basato su UAV proposto da Liu e Szirányi \cite{liu2021gesture} dal punto di vista dell’indipendenza dei soggetti. Analizzando la matrice di confusione pubblicata, le curve di apprendimento riportate e la composizione del dataset, abbiamo identificato diversi indicatori compatibili con una non indipendenza tra gli insiemi di addestramento e di test. In particolare, la combinazione di un numero ridotto di partecipanti registrati e di una suddivisione casuale a livello di campioni rende molto probabile la presenza degli stessi individui in entrambi gli insiemi, impedendo una valutazione affidabile della generalizzazione verso persone mai osservate in precedenza.

Le caratteristiche prestazionali osservate — accuratezza di classificazione quasi perfetta, matrice di confusione quasi completamente diagonale e curve di apprendimento di training e test fortemente sincronizzate — sono difficili da conciliare con il comportamento tipico dei compiti reali di riconoscimento dei gesti caratterizzati da variabilità umana. Piuttosto che costituire una prova diretta di una reale comprensione dei gesti, tali risultati sono più plausibilmente spiegati dalla presenza di informazioni specifiche dei partecipanti condivise tra le partizioni dei dati.

Lo scopo di questa analisi non è valutare la qualità implementativa del metodo originale, bensì evidenziare un problema metodologico più generale nella ricerca di apprendimento automatico centrata sull’uomo. Quando i dataset contengono misurazioni ripetute degli stessi individui, una suddivisione casuale a livello di frame o di campioni può portare a stime delle prestazioni eccessivamente ottimistiche. Per applicazioni come le operazioni di soccorso assistite da UAV, nelle quali i sistemi devono interpretare gesti eseguiti da persone sconosciute, una valutazione indipendente dai soggetti è quindi essenziale.

Più in generale, i dataset destinati alla ricerca sul riconoscimento dei gesti dovrebbero essere progettati in modo da riflettere le condizioni di impiego reale. I set di addestramento e di test dovrebbero contenere partecipanti diversi e, quando possibile, anche ambienti e sfondi differenti, affinché i modelli non possano basarsi su indizi specifici delle persone o delle scene. Ciò è particolarmente importante negli scenari di soccorso, in cui un UAV deve riconoscere gesti eseguiti da individui sconosciuti in condizioni variabili di abbigliamento, illuminazione e ambiente. La progettazione di dataset con diversità di soggetti e di contesti costituisce quindi un prerequisito necessario per valutare la capacità di generalizzazione nel mondo reale.

\section{Resumen}
Este trabajo examina el protocolo de evaluación del sistema de reconocimiento de gestos basado en UAV propuesto por Liu y Szirányi \cite{liu2021gesture} desde la perspectiva de la independencia entre sujetos. Mediante el análisis de la matriz de confusión publicada, las curvas de aprendizaje reportadas y la composición del conjunto de datos, identificamos varios indicios compatibles con una falta de independencia entre los conjuntos de entrenamiento y prueba. En particular, la combinación de un número reducido de participantes registrados y una división aleatoria a nivel de muestras hace muy probable la presencia de los mismos individuos en ambos conjuntos, lo que impide evaluar de manera fiable la generalización a personas previamente no observadas.

Las características de rendimiento observadas —una precisión de clasificación casi perfecta, una matriz de confusión prácticamente diagonal y curvas de aprendizaje de entrenamiento y prueba fuertemente sincronizadas— son difíciles de conciliar con el comportamiento típico de tareas reales de reconocimiento de gestos con variabilidad humana. En lugar de constituir evidencia directa de una comprensión genuina de los gestos, estos resultados se explican más plausiblemente por la presencia de información específica de los participantes compartida entre las particiones de datos.

El propósito de este análisis no es evaluar la calidad de implementación del método original, sino señalar un problema metodológico más general en la investigación de aprendizaje automático centrada en el ser humano. Cuando los conjuntos de datos contienen mediciones repetidas de los mismos individuos, una división aleatoria a nivel de fotogramas o muestras puede conducir a estimaciones de rendimiento excesivamente optimistas. Para aplicaciones como las operaciones de rescate asistidas por UAV, en las que los sistemas deben interpretar gestos realizados por personas desconocidas, la evaluación independiente por sujetos es, por tanto, esencial.

De manera más amplia, los conjuntos de datos destinados a la investigación en reconocimiento de gestos deberían diseñarse para reflejar las condiciones reales de despliegue. Los conjuntos de entrenamiento y prueba deben contener participantes diferentes y, cuando sea posible, también entornos y fondos distintos, de modo que los modelos no puedan basarse en pistas específicas de las personas o de las escenas. Esto es especialmente importante en escenarios de rescate, donde un UAV debe reconocer gestos realizados por individuos desconocidos bajo condiciones variables de vestimenta, iluminación y entorno. Diseñar bases de datos con diversidad de sujetos y de contextos es, por tanto, un requisito necesario para evaluar la capacidad de generalización en el mundo real.

\section{Összefoglalás}
Ez a munka a Liu és Szirányi \cite{liu2021gesture} által javasolt UAV-alapú gesztusfelismerő rendszer értékelési protokollját vizsgálja a személyfüggetlenség szempontjából. A publikált konfúziós mátrix, a közölt tanulási görbék és az adatbázis összetételének elemzése alapján több olyan jelenséget azonosítottunk, amelyek nem független tanító- és tesztfelosztásra utalnak. Különösen a rögzített résztvevők alacsony száma és a mintaszintű véletlen felosztás együttesen nagy valószínűséggel azt eredményezi, hogy ugyanazok a személyek mindkét halmazban megjelennek, ami nem teszi lehetővé a korábban nem látott emberekre való általánosítás megbízható értékelését.

A megfigyelt teljesítményjellemzők — a közel tökéletes osztályozási pontosság, a szinte teljesen diagonális konfúziós mátrix, valamint az erősen szinkronizált tanító- és teszt tanulási görbék — nehezen egyeztethetők össze a valódi, emberi variabilitást tartalmazó gesztusfelismerési feladatok tipikus viselkedésével. Ezek az eredmények nem a gesztusok valódi felismerésének közvetlen bizonyítékai, hanem sokkal inkább azzal magyarázhatók, hogy a résztvevőkre jellemző információk mindkét adatpartícióban jelen vannak.

Az elemzés célja nem az eredeti módszer implementációs minőségének megítélése, hanem egy általánosabb módszertani probléma kiemelése az emberközpontú gépi tanulási kutatásokban. Amikor az adatbázisok ugyanazon személyek ismételt méréseit tartalmazzák, a képkocka- vagy mintaszintű véletlen felosztás túlzottan optimista teljesítménybecslésekhez vezethet. Olyan alkalmazások esetében, mint az UAV-val támogatott mentési feladatok, ahol a rendszernek ismeretlen emberek gesztusait kell értelmeznie, ezért elengedhetetlen a személyfüggetlen kiértékelés.

Általánosabban, a gesztusfelismerési kutatásra szánt adatbázisokat úgy kell megtervezni, hogy tükrözzék a valós alkalmazási környezetet. A tanító- és teszthalmazoknak különböző résztvevőket kell tartalmazniuk, és lehetőség szerint eltérő környezeteket és háttereket is, hogy a modellek ne támaszkodhassanak személy- vagy jelenetspecifikus jegyekre. Ez különösen fontos mentési helyzetekben, ahol egy UAV-nak ismeretlen emberek gesztusait kell felismernie változó ruházat, megvilágítás és környezeti feltételek mellett. A személy- és kontextusdiverzitást biztosító adatbázisok kialakítása ezért a valós általánosítási képesség értékelésének szükséges feltétele.

\section{Summarium}
Hoc opus protocollum aestimationis systematis ad gestus agnoscendos ope UAV a Liu et
Szirányi \cite{liu2021gesture} propositi ex perspectiva independentiae personarum examinat. Ex analysi matricis confusionis editae, curvarum institutionis relatarum atque compositionis datorum indicia plura invenimus quae cum partitione non independenti inter copias institutionis et probationis congruunt. Praesertim coniunctio parvi numeri participantium et divisionis fortuitae ad gradum exemplorum efficit ut eadem individua in utraque copia verisimiliter appareant, quo fit ut generalizatio ad homines antea non visos fideli modo aestimari non possit.

Notae qualitatis observatae — fere perfecta accuratio classificationis, matrix confusionis paene omnino diagonalis et valde synchronae curvae institutionis atque probationis — cum typico usu recognitionis gestuum in condicionibus realibus, ubi variabilitas humana adest, difficulter conciliantur. Talia eventa non tamquam argumentum directum verae cognitionis gestuum haberi possunt, sed potius per praesentiam informationis ipsis personis propriae inter partitiones datorum communicatae explicantur.

Propositum huius analysis non est qualitatem implementationis methodi originalis iudicare, sed quaestionem methodologicam latiorem in investigatione machinali ad hominem spectante demonstrare. Cum bases datorum mensuras iteratas eorundem hominum contineant, divisio fortuita ad gradum imaginum vel exemplorum ad aestimationes qualitatis nimis optimisticas ducere potest. In applicationibus autem quales sunt operationes auxilii ope UAV, ubi systema gestus hominum ignotorum interpretari debet, aestimatio a personis independens necessaria est.

Generalius, bases datorum ad investigationem recognitionis gestuum destinatae ita constitui debent ut condiciones usus reales repraesentent. Copiae institutionis et probationis diversos participes continere debent et, si fieri potest, etiam diversa loca atque scaenas, ne exemplaria indiciis personarum vel locorum propriis niti possint. Hoc praesertim in casibus auxilii magni momenti est, ubi UAV gestus hominum ignotorum sub variis condicionibus vestitus, illuminationis et ambitus agnoscere debet. Constitutio basium datorum cum diversitate personarum et contextuum igitur praerequisitum necessarium est ad facultatem generalizationis in mundo reali recte aestimandam.

\section{Resumo}
Ĉi tiu laboro ekzamenas la taksadan protokolon de la UAV-bazita gestrekona sistemo proponita de Liu kaj Szirányi \cite{liu2021gesture} el la vidpunkto de sendependeco inter personoj. Per analizo de la publikigita konfuzomatrico, la raportitaj lernokurboj kaj la konsisto de la datumaro, ni identigis plurajn indikojn kongruajn kun ne-sendependa disigo inter la trejna kaj la testa aroj. Aparte, la kombino de malgranda nombro da registritaj partoprenantoj kaj hazarda disdivido je la nivelo de unuopaj ekzemploj faras tre verŝajna la ĉeeston de la samaj individuoj en ambaŭ aroj, kio ne ebligas fidinde taksi ĝeneraligon al antaŭe neviditaj personoj.

La observitaj rendimentaj trajtoj — preskaŭ perfekta klasifika ĝusteco, preskaŭ tute diagonala konfuzomatrico kaj forte sinkronigitaj lernokurboj por trejnado kaj testado — malfacile kongruas kun la tipa konduto de realmondaj gestrekonaj taskoj kun homa variemo. Prefere ol esti rekta pruvo de vera kompreno de gestoj, tiaj rezultoj estas pli verŝajne klarigeblaj per la ĉeesto de person-specifaj informoj dividitaj inter la datumaj partoj.

La celo de ĉi tiu analizo ne estas taksi la kvaliton de la efektivigo de la origina metodo, sed montri pli ĝeneralan metodikan problemon en hom-centra maŝinlernada esplorado. Kiam datumoj enhavas ripetajn mezurojn de la samaj individuoj, hazarda disdivido je la nivelo de bildoj aŭ ekzemploj povas konduki al tro optimismaj rendimentaj taksoj. Por aplikoj kiel UAV-subtenataj savoperacioj, kie la sistemo devas interpreti gestojn de nekonataj homoj, sendependa taksado laŭ personoj estas do esenca.

Pli ĝenerale, datumoj destinitaj por gestrekona esplorado devus esti desegnitaj tiel, ke ili reflektu realajn uzkondiĉojn. La trejnaj kaj testaj aroj devus enhavi malsamajn partoprenantojn kaj, kiam eble, ankaŭ malsamajn mediojn kaj fonojn, por ke modeloj ne povu dependi de person- aŭ scen-specifaj indikoj. Tio estas aparte grava en savoscenaroj, kie UAV devas rekoni gestojn faritajn de nekonataj individuoj sub ŝanĝiĝantaj kondiĉoj de vestaĵo, lumo kaj medio. Projektado de datumoj kun diverseco de personoj kaj kuntekstoj estas tial necesa antaŭkondiĉo por ĝuste taksi realmondan ĝeneraligan kapablon.

\section{Summario}
Iste labor examina le protocollo de evaluation del systema de recognition de gestos basate sur UAV proponite per Liu e Szirányi \cite{liu2021gesture} ab le perspectiva del independentia del subjectos. Per le analyse del matrice de confusion publicate, del curvas de apprendimento reportate e del composition del base de datos, nos ha identificate plure indicios compatibile con un partition non independente inter le insimules de entrainamento e de test. In particular, le combination de un numero reducite de participantes registrate e un division aleatori al nivello de exemplares rende multo probabile le presentia del mesme individuos in ambe insimules, lo que non permitte evaluar de maniera fiabile le generalisation a personas previemente non observate.

Le characteristicas de rendimento observate — un precision de classification quasi perfecte, un matrice de confusion quasi completemente diagonal e curvas de apprendimento de entrainamento e test fortemente synchronisate — es difficile de reconciliar con le comportamento typic de cargas real de recognition de gestos con variabilitate human. In loco de constituer evidentia directe de un ver recognition de gestos, tal resultatos es plus plausibilemente explicate per le presentia de information specific al participantes compartite inter le partitiones del datos.

Le objectivo de iste analyse non es evaluar le qualitate implementational del methodo original, sed signalar un problema methodological plus general in le recerca de apprendimento automatic centrate sur le humano. Quando le bases de datos contine mesuras repetite del mesme individuos, un division aleatori al nivello de imagines o exemplares pote conducer a estimationes de rendimento troppo optimista. Pro applicationes como operationes de salvamento assistite per UAV, ubi le systema debe interpretar gestos de personas incognite, un evaluation independente per subjectos es dunque essential.

Plus generalemente, le bases de datos destinate al recerca de recognition de gestos deberea esser projectate pro reflecter le conditiones real de uso. Le insimules de entrainamento e de test deberea continer participantes differente e, quando possibile, etiam ambientes e fundos differente, pro evitar que le modellos se basa sur indicios specific al personas o al scenas. Isto es particularmente importante in scenarios de salvamento, ubi un UAV debe recognoscer gestos effectuate per individuos incognite sub conditiones variabile de vestimento, illumination e ambiente. Le conception de bases de datos con diversitate de subjectos e de contextos es dunque un prerequisito necessari pro evaluar correctemente le capacitate de generalisation in le mundo real.
\bibliographystyle{unsrt}  
\bibliography{references}  






\end{document}